# ChatGPT in the context of precision agriculture data analytics

Ilyas Potamitis,[1]

[1] Department of Music Technology and Acoustics, Hellenic Mediterranean University, 74100 Rethymno, Greece.

Correspondence should be addressed to Prof. Ilyas Potamitis; potamitis@hmu.gr

## Abstract

In this study we argue that integrating ChatGPT into the data processing pipeline of automated sensors in precision agriculture has the potential to bring several benefits and enhance various aspects of modern farming practices. Policy makers often face a barrier when they need to get informed about the situation in vast agricultural fields to reach to decisions. They depend on the close collaboration between agricultural experts in the field, data analysts, and technology providers to create interdisciplinary teams that cannot always be secured on demand or establish effective communication across these diverse domains to respond in real-time. In this work we argue that the speech recognition input modality of ChatGPT provides a more intuitive and natural way for policy makers to interact with the database of the server of an agricultural data processing system to which a large, dispersed network of automated insect traps and sensors' probes reports. The large language models map the speech input to text, allowing the user to form its own version of unconstrained verbal query, raising the barrier of having to learn and adapt oneself to a specific data analytics software. The output of the language model can interact through Python code and Pandas with the entire database, visualize the results and use speech synthesis to engage the user in an iterative and refining discussion related to the data. We show three ways of how ChatGPT can interact with the database of the remote server to which a dispersed network of different modalities (optical counters, vibration recordings, pictures, and video), report. We examine the potential and the validity of the response of ChatGPT in analysing, and interpreting agricultural data, providing real-time insights and recommendations to stakeholders.

## Introduction

The ever-growing global population places unprecedented demands on agriculture to increase food production while ensuring environmental sustainability. Climate change and invasive insects apply additional pressure, challenging the adaptability and resilience of agricultural ecosystems.
It is estimated that insect pests damage 18–20% of the world's annual crop production, which is worth more than USD 470 billion [1]. Fruit fly and moth pests inflict staggering economic damage on agriculture, with substantial monetary losses documented globally. For instance, fruit flies are responsible for annual losses estimated at billions USD, affecting a variety of fruits and vegetables. The Mediterranean fruit fly alone (*Ceratitis capitata*) accounts for billions of dollars in damage to citrus crops in the United States. Similarly, moth pests, particularly the cotton bollworm (*Helicoverpa armigera*), are known to cause extensive harm, with annual losses exceeding $5 billion USD in various crops, including cotton, maize, and soybean [2-6]. The tomato leaf miner (*Tuta absoluta*) is another economically damaging moth pest, causing losses ranging from 11% to 43% annually, with potential devastation reaching

100% if control measures are insufficient. These economic figures [7-10]. underscore the urgent need for effective pest management strategies and innovative monitoring technologies to safeguard agricultural production and global food security.

Other pests like wood-boring insects, including termites, bark beetles, and a range of Cerambycidae Coleoptera, present formidable challenges to the economy. Wood- and phloem-boring insects are inducing nearly $1.7 billion in local government expenditures and approximately $830 million in lost residential property values annually [11]. Bark beetles, particularly the mountain pine beetle, have led to timber losses amounting to millions of dollars, not to mention the cascading impacts on related industries like tourism and increased forest fire risks. Additionally, powderpost beetles can severely compromise the structural integrity of lumber in buildings and furniture.

Effective control measures require timely interventions that can only be guaranteed if a pest population monitoring protocol is in effect from the beginning till the end of crop season. These protocols try to assess where, when and at what stage the problem has evolved. Monitoring of pests is usually carried out by various paper or plastic traps, but manual monitoring cannot be performed at a large scale, spatially and temporally, due to manpower and cost constraints. To relieve these constraints, automated insect traps have emerged as powerful tools in precision agriculture, offering various ways to monitor insect populations. Automated insect traps are equipped with sensors and wireless communication capabilities to detect, count and sometimes, identify the targeted insects, providing counts of sampled insect fauna in specific areas of a field. Automated traps can provide early warning systems for the presence of invasive or damaging pests, allowing farmers to take proactive measures to protect their crops before significant damage occurs. Data from automated traps can be integrated into decision support systems to help farmers make informed decisions about when and where to apply pesticides or employ alternative pest control methods, reducing the overall use of chemicals. This early detection can be crucial in preventing the establishment and spread of invasive species, potentially saving entire crops or ecosystems from devastation. Finally, automated traps provide real-time data on insect activity, which helps farmers optimize the timing of interventions like pesticide application, irrigation, or planting and prioritize interventions and allocate resources more efficiently, reducing waste and environmental impact.

There are mainly three major approaches to automated insect monitoring, each with unique mechanisms and applications.

(a) **Optical Counters in the entrance of traps:** These devices are attached to the entrances of fruit fly and moth traps and primarily target specific pests using species-specific lures. The optical counters are reliant on the specificity of these lures to attract the targeted insects. This approach ensures that the traps are selective, only attracting and counting specific insect species or groups [12-17].

(b) **Camera-Based traps:** In this approach, traps are equipped with cameras that capture images of their internal space, allowing for the visual documentation of trapped insects. These images are then analysed, manually or automatically using machine learning techniques, to identify the various insect species present in the trap. There is a rich body of research and various approaches that involve embedding cameras in insect traps, enhancing the effectiveness and precision of insect monitoring and identification processes [18-27].

(c) **Near-Infrared sensors and Lidars:** This technology utilizes infrared and near-infrared sensors and lidars that emit light, covering a volume of space in the open field. These sensors register the backscattered wingbeat signals of flying insects, allowing for the detection of flying insect fauna and assessment of its activity within a specific area. Unlike the other methods, this approach allows for the monitoring of insects in a designated volume of space, providing real-time insights into insect activity and behaviour [28-30]. Each of these automated insect monitoring approaches has its advantages and specific use cases, contributing to the advancement of pest monitoring and management strategies.

An extensive body of literature on these technologies, comprising hundreds of published papers enhances our understanding of insect populations, behaviours, and the impact they have on various ecosystems and agricultural practices. In this work, we predominantly reference review papers, providing a curated overview of applied methodologies without detailed exploration and evaluation of their contributions. *The interested reader needs to be aware that in this document, we undertake a comprehensive review of our approach on automated insect surveillance leading to a presentation that is inherently oriented towards our methodology within the context of existing research.* Our vision is to transform typical plastic traps to cyber-physical objects that sense and count/identify the captured insect and transmit all information using internet of things (IoT) communications. Sensing, counting, and identification of captured insects involves the use of artificial intelligence, and we are interested in all insect-cases of economic importance with a view to deploy e-traps/probes at global scales (see Fig. 1). All these wireless networks share a common server that holds the database of all sensing modalities.

In this work, we argue that large language models (LLM) like ChatGPT can become the new disruptive interface that interacts via unconstrained speech with large networks of automated modalities that monitor insects of economic importance, extending around the globe and we demonstrate this through the remote server of Insectronics [31]. This server hosts a database to which an expanding network of hundreds of automated devices located in several countries reports. Namely, a) e-traps of agricultural lepidoptera, b) probes that transmit recordings of the micro-vibrations produced by woodboring insects, and c) wingbeat recorders of flying insects and vision-based traps for insects of economic importance. This report used the ChatGPT but in principle any LLM is compliant to the core idea. The idea is to use voice input as a more intuitive and natural way to interact with the agricultural data processing system without the need for complex platforms or manual data entry. Once the speech input is transcribed to text by the speech recognition module, the LLM will map to text the questions and commands that are expressed in diverse ways to the same queries. The queries are then mapped into Python code (in our case either with ChatGPT4-PLUS or OpenAI's LangChain [32] controlling the Pandas DataFrame) that accesses the database and makes queries and inferences, executes commands and visualizes the response while giving evidence of its algorithmic approach through text and the provision of the Python code. Lastly, the outcome of the query can be announced using seamlessly cooperating speech synthesis software. ChatGPT and related LLMs can engage into an iterative and refining dialog until the information is extracted for a holistic approach in data extraction.

## Materials and Methods

Insectronics has provided us with partial access to their database and their application programming interface (API) from where collective results and examples can be drawn. A short description of the devices that have been reporting to Insectronics server follows:

**The e-funnel trap**

The attractiveness of the typical funnel trap is established and has been used in the field for a long time. The e-funnel (Fig. 2) is a typical plastic funnel trap with an add-on optical counter that counts moths as they fall into the trap and interrupt the flow of light from emitter to receiver.

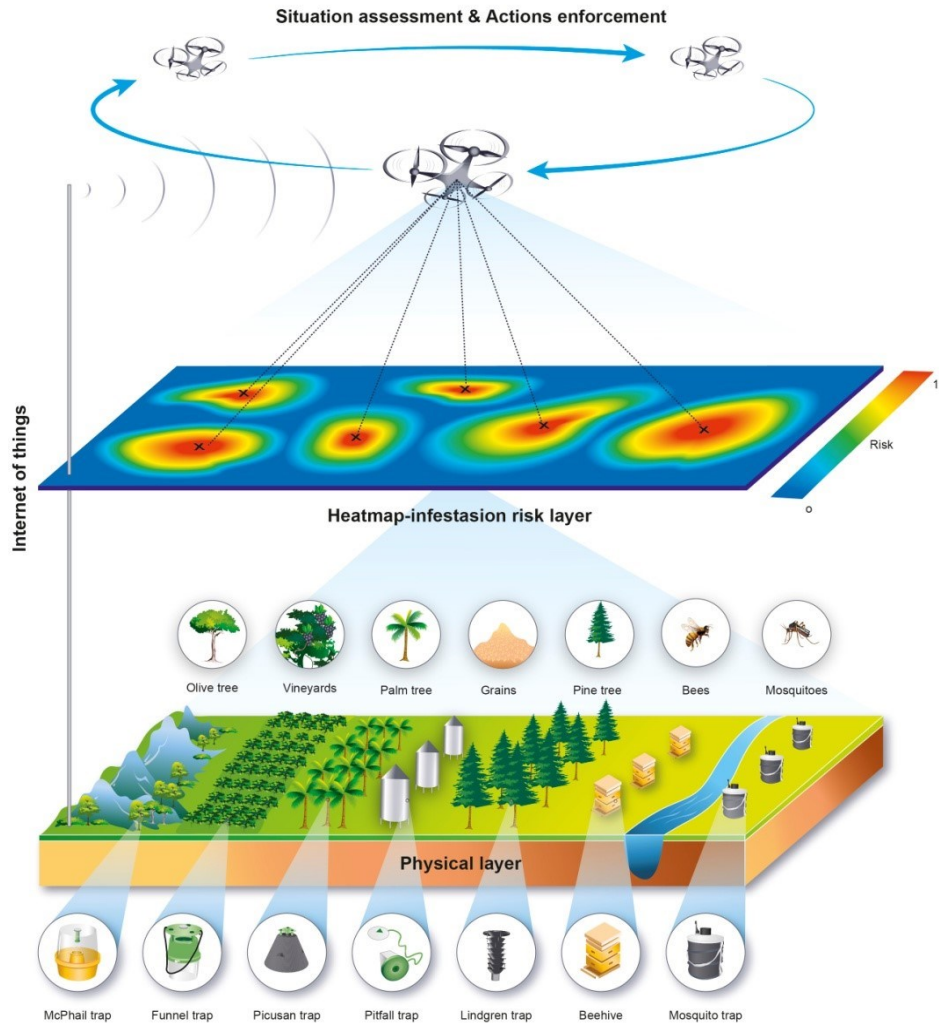

Figure 1. We envisage optical counters embedded in typical insect traps that report through IoT to a common server. The server may direct drones for interventions automatically.

The embedded modem collectively uploads insect counts, timestamps of captured insects, GPS coordinates and environmental parameters through the mobile network to the cloud server. The communication of the network with the cloud server is bi-directional: trap data are uploaded to the server and the e-traps can be reconfigured by receiving instructions from the server (on-off commands, time scheduling, tuning of thresholds related to insect detection, local time-zones etc.). It is an instance of an internet of things (IoT) application in agriculture. The electronics of the e-traps are low power making them power-sufficient for the whole monitoring period without solar panel or recharging. They can hold slow-release pheromone dispensers that last for many months diminishing the need for maintenance. Lastly, because in field situations we may be facing high pest pressure if the number of daily catches is very high (e.g., 200-300/day), the bucket can self-dispose the captured insects [33] or have a net that elongates its capture capacity and does not face the problem of insects piling up and filling the trap (Figure 2-Left). They can avoid double counts of insects and the effect of harsh weather conditions. The e-funnel can reliably detect the onset of the infestation, when and where the infestation goes out of bounds and assess the effect of a treatment by monitoring the number of aftertreatment insect counts because it transmits 4 times per day (this should be contrasted with manual visiting every 7-10 days). Because they are maintenance free and autonomous, monitoring agencies can expand their contracts at an unprecedented number and deploy a large number of nodes extending in many countries. A proof-of-concept study of the e-funnel is presented in [14]. They monitor harmful pests of Lepidoptera species of major crops (olive, vines, pear, chestnut, apple etc) depending on the pheromone used, such as *Helicoverpa armigera*, *Pectinophora gossypiella*, *Prays oleae* (Bernard) (Lepidoptera; Yponomeutidae), *T. absoluta*, E. bigella, *Thaumetopoea pityocampa* (Denis & Schiffermüller) (Lepidoptera: Thaumetopoeidae), *Lobesia botrana* (Denis & Schiffermüller) (Lepidoptera: Tortricidae), *Cydia splendana* (Hobner) (Lepidoptera:Tortricidae), *Cydia pomonella* in apple orchards.

The structure of the e-funnel's tabular data is seen in Table 1:

| Timestamp | Counts | Temperature | Humidity | Lat | Long | Name |
|---|---|---|---|---|---|---|
| 01-06-23 8:00 | 8 | 24.61 | 66.2 | 39.30149 | 22.33027 | 100 |
| 01-06-23 9:00 | 6 | 26.99 | 57.4 | 39.30149 | 22.33027 | 100 |
| 01-06-23 10:00 | 4 | 30.2 | 48.8 | 39.30149 | 22.33027 | 100 |
| 01-06-23 11:00 | 12 | 31.57 | 43.2 | 39.30149 | 22.33027 | 100 |

Table 1. Counts refer to insect counts. Humidity (%) and Temperature (ºC) are hourly measurements. Latitude (Lat) and longitude (Long) refer to the GPS coordinates of the devices. 'Name' refers to an identification index unique for each device.

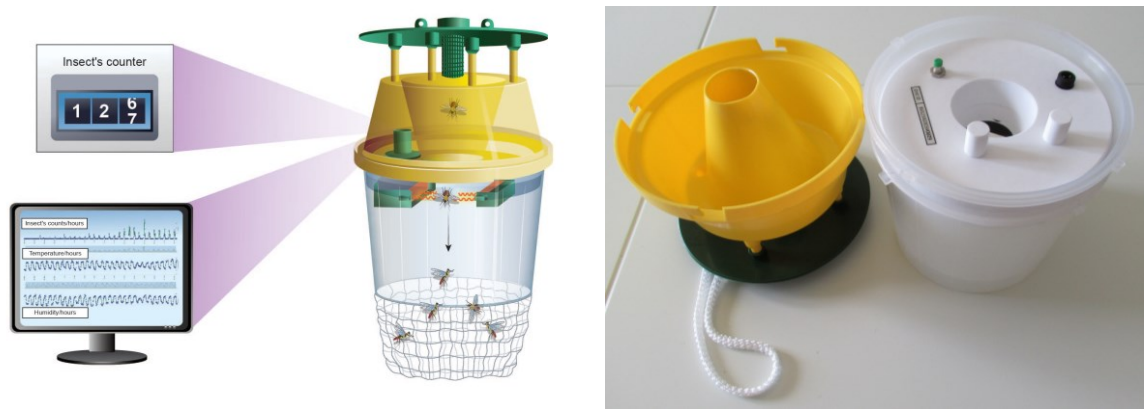

Figure 2. (Left) The e-funnel trap. (Right) The optical counter attached to the funnel trap.

**The TreeVibe probe for monitoring woodboring insects**

The intended use of the device is the long-term systematic monitoring of trees and wooden structures against borers. This device is composed of a) a waveguide, b) a piezolelectric sensing element and c) the electronic board that transforms recorded vibrations to lineout audio signals that are stored, compressed, and wirelessly transmitted (see Fig. 3). The waveguide is just a stainless-steel bar that serves as a sound coupler between the wood and the sensing element and in order not to harm the tree can become long or very thin and short depending on the diameter of the tree trunk. The locomotion and feeding of borers inside the tree or under its bark cause the propagation of vibrations that are picked up by the waveguide and are led to the stable piezo-ceramic crystal that slightly deforms, and as a result, outputs voltage variations. These voltage variations are amplified and processed by low-power electronics board. The device carries a GPS and a communications modem to transmit the recordings, environmental variables, and the device's position. The device samples periodically the internal soundscape of the tree (e.g., every hour) and falls to sleep. At some point in time, it wakes up transmits collectively the recordings to the server and goes back to sleep. The parameters of the recording and reporting sessions are handled by the server, which also holds the transmitted recordings in its database. The user can listen to the recording remotely, process them automatically to infer the infestation state of the tree with wood-boring insects that feed or move inside the tree. It has been extensively in field trials of the pests *Xylotrechus chinensis, Rhynchophorus ferrugineus* (Red palm weevil)*,* mountain bark beetle*,* emerald ash borer*, Rhynchophorus palmarum, Aromia bungii* (Red necked longicorn)*, Anoplophora glabripennis* (Asian longhorn beetle) *and Anoplophora chinensis* (Citrus longhorn beetle). The early idea of vibroacoustic monitoring of trees in not new and is presented in [34] among other references. The remote version in the form of the TreeVibe device is presented in [35] and [36].

The structure of the TreeVibes records is a list of files like the following: F_20230812193118_1.mp3. F stands for File, 2023 stands for the year, 08 for the month (August), 12 for the day of the month, 19 for the hour of the day, 31 for the minute of the hour, 18 for the second of the minute (therefore the time stamp is 12/08/2023 19:31:18). The symbol _1 is a serial number of the recording.

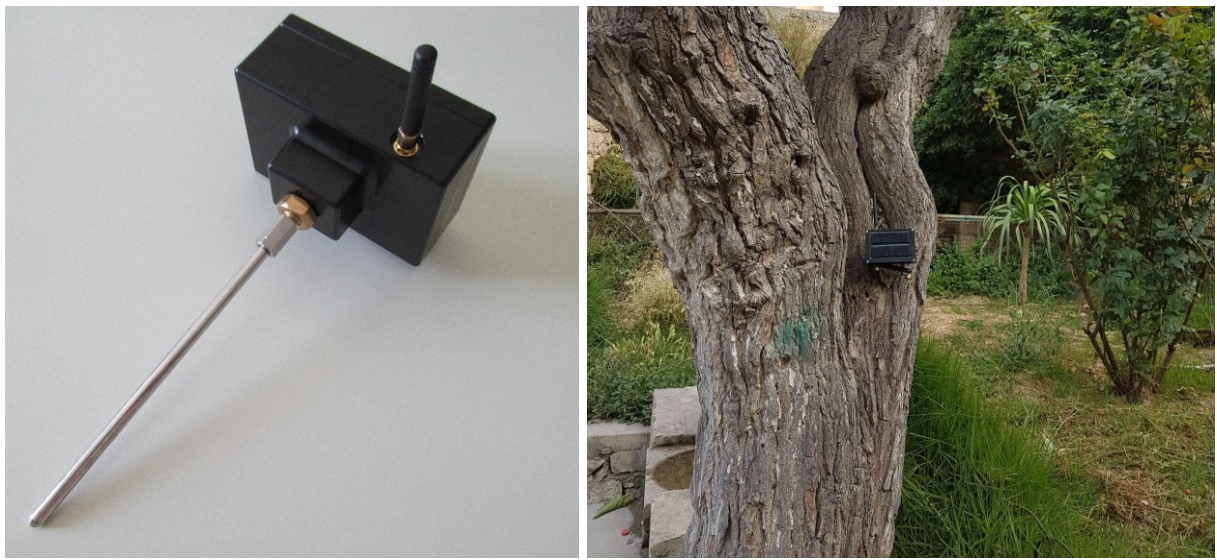

Figure 3. (Left) The main enclosure contains the electronics. We can see a communication antenna on top and a metal waveguide acting as a vibration's coupler. The device has in its back an embedded solar panel connected to its battery. (Right) The device attached to a tree listening for wood boring insects.

**The Wingbeats recorder**

The basis of the insects' wingbeats recorder has been reported in [33-36] for extinction light and in [37] for backscattered light. We report results based on the former one. The central idea is that a flow of infrared light from emitter to receiver (see Fig. 4-Left) is interrupted and modulated by flying insects inside its field of view (FOV). This modulation results in light fluctuations due to the casted shadow in the output of the receiver. The light fluctuations are turned to voltage variations by the receiver's photodiodes. The receiver is monitored by a microprocessor and the device is auto triggered when it senses a wingbeat. Once triggered, the device registers a continuous recording of light fluctuations that is normalized between [-1, 1] and is turned to an audio snippet, that is time-stamped and internally stored in the SD of the device. The emitting and receiving aperture of the sensor is made large enough to accommodate the full motion of a wingbeat by Fresnel lenses that expand the beam and collimate light to allow tracking of fast flying insects such as fruit flies that would otherwise spend little time inside the FOV. It has been extensively used to record the wingbeat of various insects including the wingbeat of Culicoides [37], bees [38], mosquitoes [39], fruit flies [40-41] and aphids [42], (see Fig. 4-Right for a configuration that allows this). The device has an ethernet connection and uploads the recordings in real time from the different parts of the world that has been deployed. Related research work that leads to wingbeat recorder embedded in fruit fly traps have been presented in [43-45].

The structure of the Wingbeats recordings is a list of files like the following:

F_20230430001523_0000_Temp22.5_Hum54.9_Opt00.00.wav

F stands for File, 2023 stands for the year, 04 for the month (April), 30 for the day of the month, 00 for the hour of the day, 15 for the minute of the hour, 23 for the second of the minute (therefore the time stamp is 30/04/2023 00:15:23)

0000 is a serial number of the recording, Temp22.5 means temperature 22.5°C, Hum54.9 means 54.9% relative humidity, Opt00.00 means 0 optical intensity (i.e., darkness)

ChatGPT easily parses the filename and retrieves the requested data based on the timestamp.

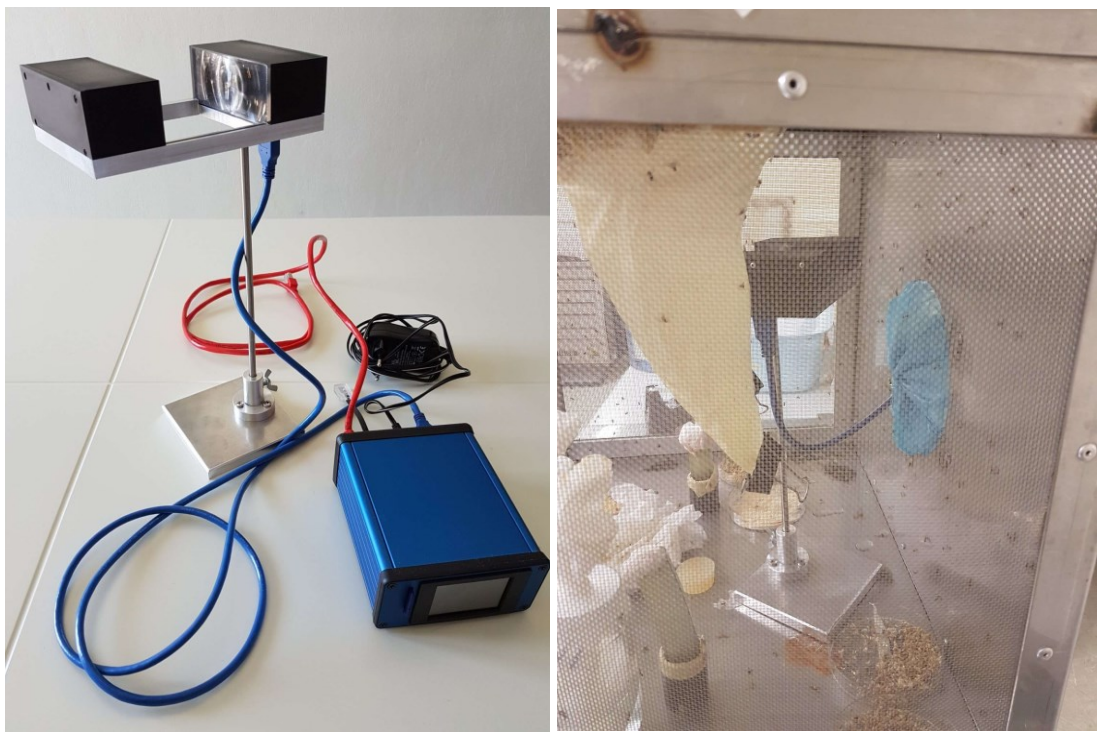

Figure 4. (Left) The wingbeat recorder and its associated optical sensor. (Right) The recorder inside a BugDorm recording the wingbeat of any flying insect crossing its field of view.

**The vision-based traps**

As a camera-based trap, in this work we use the device reported in [25]. The camera is based on top, looking down to the bottom of the bucket. The imaging sensor is integrated with a Raspberry Pi Zero microcontroller unit that controls the camera and has a YOLO-based embedded deep learning architectures that count agricultural pests by analyzing the image (see Fig. 5). The Raspberry Pi Zero is controlled by another independent microcontroller that takes on the power management of the Raspberry and completely shuts it down or powers it on to save power. The external microcontroller communicates with the Raspberry Pi through a serial bus and takes on the power management, and the calendar of all processes. The device also receives commands from the server that configures its operation whereas an embedded servomotor can automatically rotate the detached bottom of the bucket to dispose hydrated insects once they start to pile up [33].

Normally, in operational conditions, only the recognition results, insect counts, and environmental data are uploaded but for the current investigation it has been configured to upload the images to the Insectronics server. Different versions of the vision-based traps, in the context of our work, have been presented in [25-27].

The structure of the Wingbeats recordings is a list of image files like the following: F_20230908180311_0.jpg (2023-09-08 18:03:11). The filename structure is the same as the TreeVibes and Wingbeat recorder. ChatGPT parses the filename and retrieves the wanted data relying on the query information provided by the user.

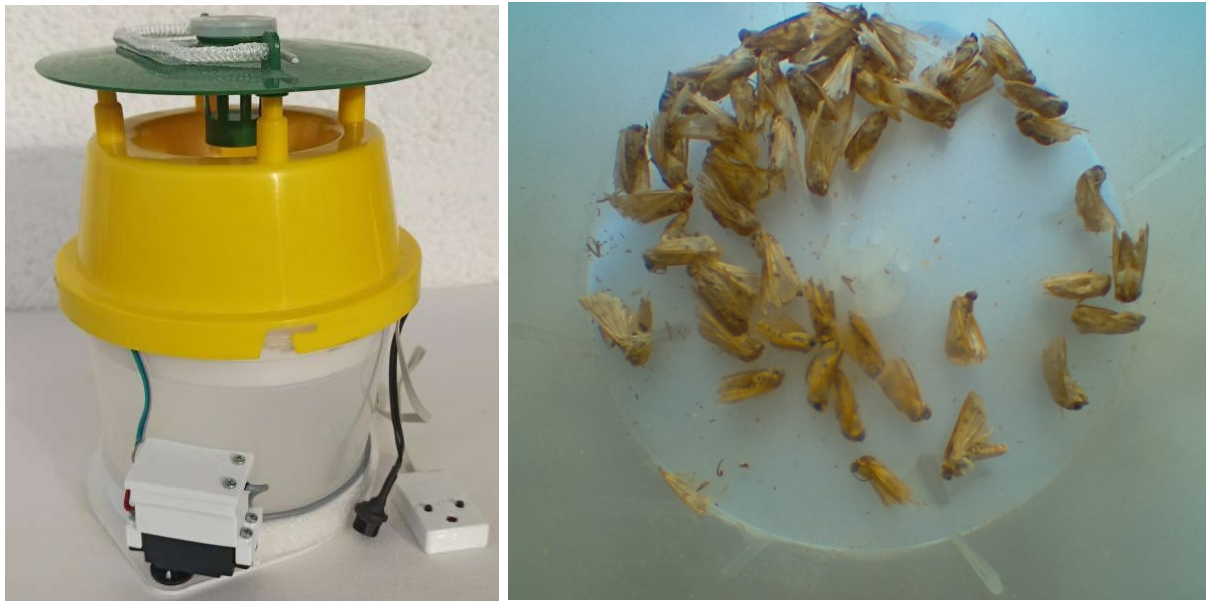

Figure 5. (Left) A vision-based trap. The camera is looking down to the bucket's floor, (Right) A picture of *H. armigera* catches transmitted from inside the bucket.

## Results and Discussion

We have identified three ways to make ChatGPT access the server's data:

a) First, one can extract all data from the database and upload a single collective csv file to the upload plugin of Advanced Data Analysis of GPT4-PLUS. Then one can use speech-to-text and text-to-speech technology embedded in the browser to allow two-way voiced interaction with ChatGPT, or type in the questions to be answered.

b) The second is to use LangChain [32] with an Api_key from OpenAI that cooperates with Pythons' Pandas DataFrame and pass the entire database to the DataFrame (see APPENDIX) or create a ChatBox that focuses on query-answering based on this specific database.

c) The third way is to allow ChatGPT to make API calls to the Insectronics server and extract all the information it needs to further process. Once extracted, the questions-answers iterations and refinement can continue as above.

For the sake of space and manuscript's clarity we hereinafter include the queries and the response (denoted as ANS) the LLM produces, and, in the APPENDIX, we include the link to the github page where one can find the Python code produced by ChatGPT. The code is cross-checked manually to produce the correct results.

We suggest to the interested reader to observe that a stakeholder interacting with the database through the ChatGPT can ask freely, using speech interaction, the following questions. One gets an analysis of the situation and conclusions without having to be a computer expert whereas until now, these answers would require hardcoded software platforms and the cooperation of data scientists. One can immediately retrieve and visualize data to grasp the situation without the need and delays of interdisciplinary teams. We note that this is a significant change in the way of interacting with a precision-agriculture related database. The stakeholder has two progressive ways to evaluate the validity of the ChatGPT's answer: First, one can examine the tactic that the LLM used to face the posed problem, as the LLM states its intention. Second, one can require seeing the Python code to examine its syntax. The LLM provides the implementation code as well and most importantly, the data visualizations. All produced and evaluated code by us can be found in the Appendix.

**The e-funnel trap**

In 2023, 250 automatic e-funnel traps have been deployd in Greece, Romania, Spain, and Italy, to monitor agricultural Lepidoptera from the beginning of the monitoring season till its end. Specifically, they predominantly monitored *H. armigera* in cotton plantations, but also *Tuta absoluta* in tomatoes, *Prays oleae* in olive trees, *Lobesia botrana* in vineyards, *Cydia pomonella* in apple orchards. In the following analysis a timeframe of data from 1/6/2023-1/9/2023 have been used. The structure of the database is in Table 1, and the Python code will cooperate with any database that has the same structure.

**Simple Queries (for all the queries the ChatGPT4-Plus prompt includes: 'Provide Python code'):**

*Q1: Which traps recorded the highest and lowest insect counts in an hour? In a day? What about in a week? Identify their location in each case.*
ANS: To answer your questions, I'll proceed with the following steps:
**Highest and Lowest Insect Counts in an Hour**:
Identify the traps that recorded the highest and lowest insect counts in any given hour, along with their locations.
**Highest and Lowest Insect Counts in a Day**:
Sum the insect counts for each trap for each day, and then identify the traps with the highest and lowest total counts, along with their locations.
**Highest and Lowest Insect Counts in a Week**:
Sum the insect counts for each trap for each week, and then identify the traps with the highest and lowest total counts, along with their locations.
```
In an Hour:
Highest Count:
Trap ID: 149
Location: Latitude: 39.243683, Longitude: 22.021027
Timestamp: 2023-07-12 20:00:00
Insect Count: 319
Lowest Count:
Trap ID: 100
Location: Latitude: 39.301493, Longitude: 22.330266
Timestamp: 2023-06-01 00:00:00
Insect Count: 0 (Note: There might be multiple instances with a count of 0;
this is just the first occurrence.)
```
<The output continues for the rest of the questions omitted for brevity>

*Q2: Is there a significant difference in insect counts between adjacent (<1 km distance) traps?*
ANS: To perform ANOVA (Analysis of Variance) to assess the differences in insect counts between adjacent traps within 1 km, we will:
**Calculate the distances** between each pair of traps and identify those that are within 1 km of each other.

**Group the data** by these identified adjacent trap pairs.
**Perform ANOVA** to test whether there is a significant difference in the insect counts among the adjacent traps.
Let's start by identifying the adjacent traps based on the 1 km distance threshold.
We have identified 25 pairs of traps that are adjacent to each other, with each pair being within 1 km of distance. The ANOVA test results show a significant difference in insect counts among the adjacent trap pairs:
F-statistic: 75.20, p-value: 0.00
Since the p-value is very small (close to zero), we can reject the null hypothesis and conclude that there is a significant difference in insect counts among the adjacent traps (within 1 km of each other).

*Q3: How does temperature vary across different traps? Pick five trap names at random and visualize this somehow and overlay a plot of min max values?*
ANS (see also Fig. 6): Here's the visualization for the temperature variation across five randomly selected traps:
The boxplots show the distribution of temperatures for each trap, where the red line indicates the median temperature. The blue points represent the minimum temperatures recorded for each trap. The green points represent the maximum temperatures recorded for each trap. This visualization provides insight into how temperature varies for these randomly chosen traps, including the range, median, and potential outliers.

*Q4: Can you somehow quantify the correlation between temperature and humidity across all traps? Visualize for 2 traps.*
ANS (see also Fig. 7): To quantify the correlation between temperature and humidity, we can use the Pearson correlation coefficient. This coefficient ranges from -1 to 1, where: r=1 implies a perfect positive correlation, r=−1 implies a perfect negative correlation, r=0 implies no correlation. For visualization, we can create scatter plots for each of the five randomly selected traps to visually inspect the relationship between temperature and humidity. The Pearson correlation coefficient will be displayed on each plot.

*Q5: Are there any trends or patterns in insect counts over time?*

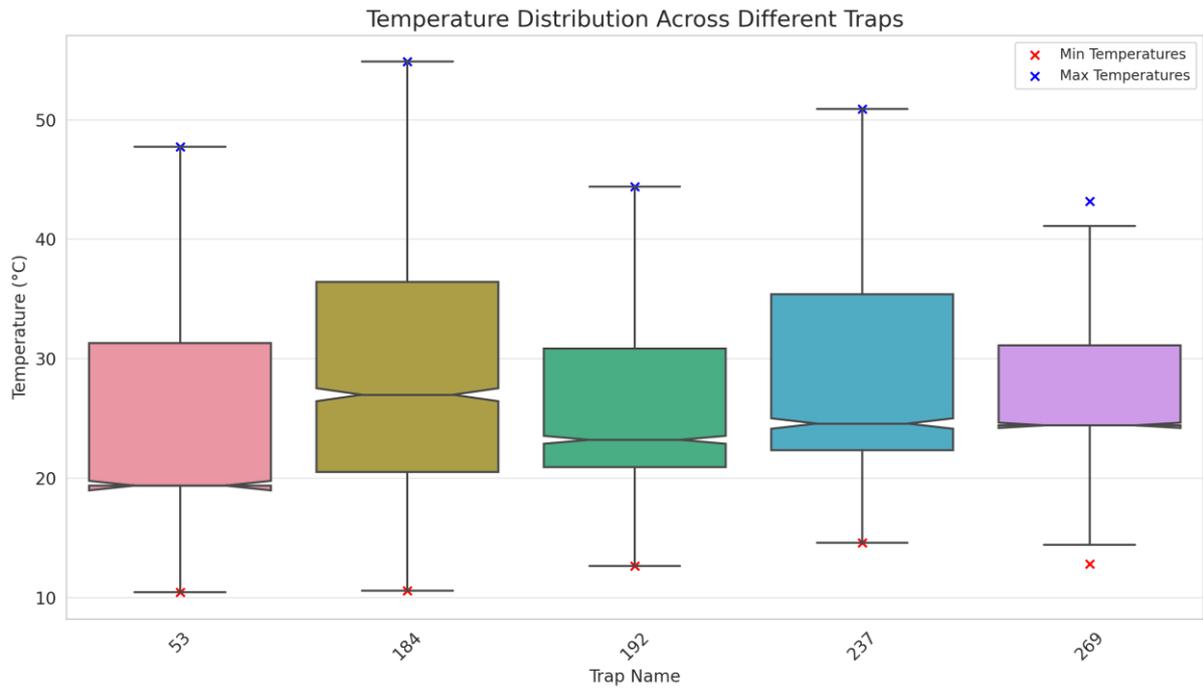

Figure 6. Box plots of temperature for 5 random e-traps with min-max temperatures superimposed. We immediately see that in the field temperatures can supersede 50 °C, whereas during summer the temperatures in these locations were mainly between 25-30 °C.

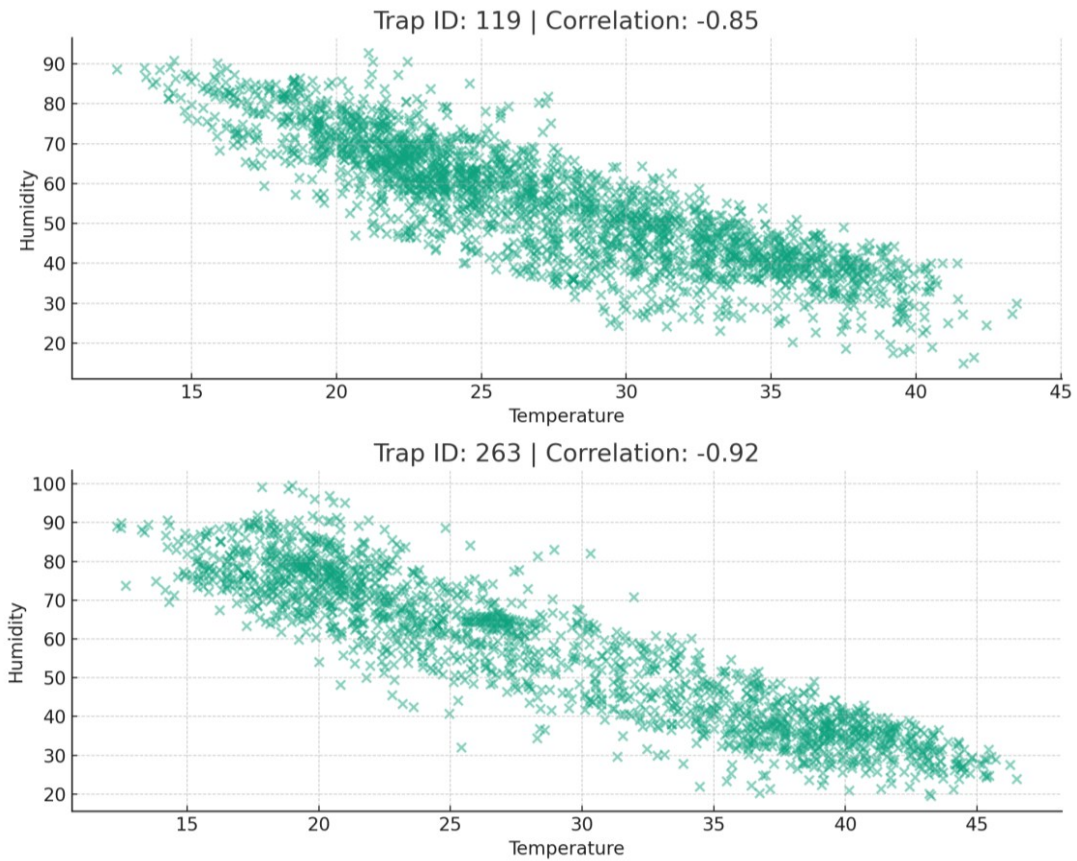

Figure 7. Scatterplots of temperature vs Humidity for 2 random e-traps. We discern the negative correlation between temperature and humidity. Temperature raises during the day and humidity increases at night.

ANS (see also Fig. 8): To analyse trends or patterns in insect counts in traps over time, we can take the following approach: Convert the 'Timestamp' column to a datetime format. Group data by trap ID ('Name') and by a time interval (e.g., daily, monthly) to aggregate insect counts. Plot the aggregated insect counts over time for a few randomly selected traps to visually inspect for trends. Optionally, use a regression or smoothing technique to highlight trends.

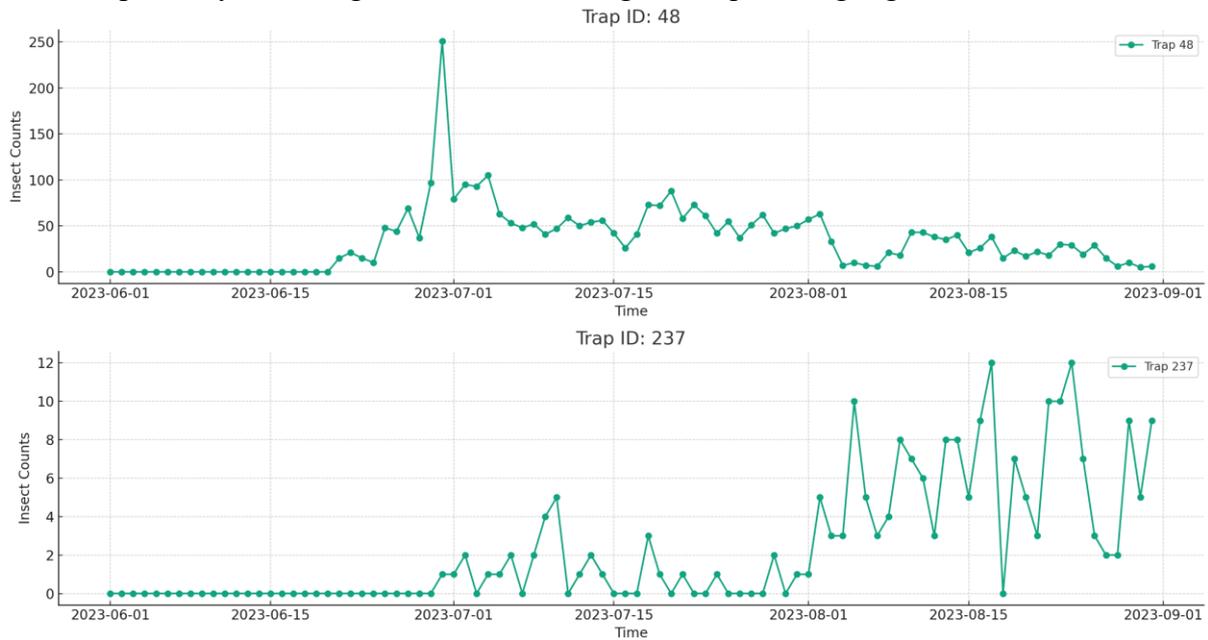

Figure 8. Timeseries of insect counts for 2 random e-traps. We see the increasing number of insects captures from June to August.

*Q6: What is the average daily count of insects for each trap? Visualize the first ten with the higher average value.*

ANS (see also Fig. 9): I'll group the data by trap name and date. Calculate the total insect count for each group. Calculate the average of these daily totals for each trap. Finally, I'll visualize the top 10 traps with the highest average daily insect counts. Let me do this for you.

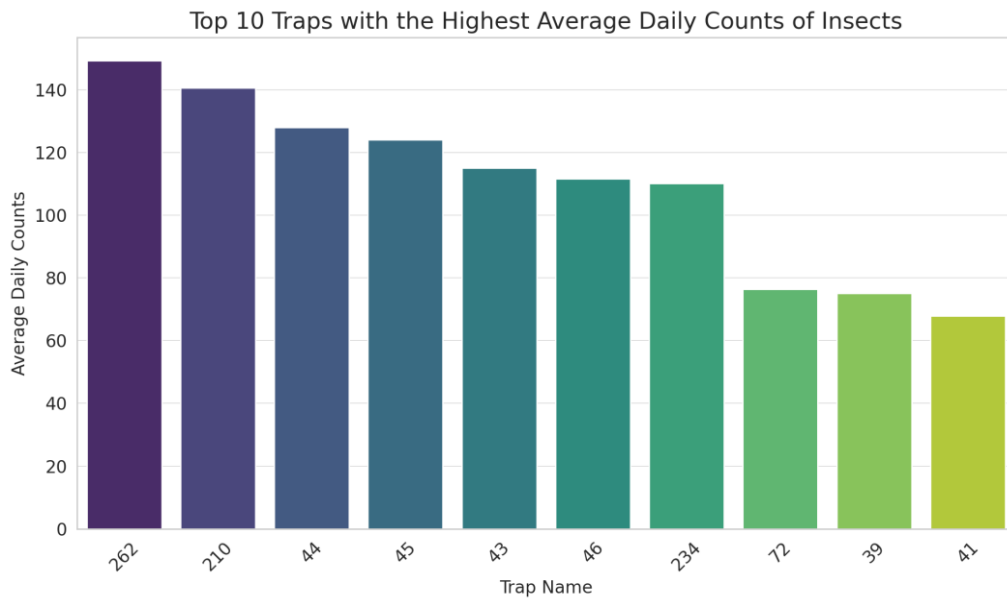

Figure 9. Bar charts of the highest daily mean captures of insects for ten e-traps.

*Q7: What time are the insects most active (circadian rhythm). Show a heatmap of activity where the y-axis is the hours of the day and x-axis the days. Present a colorbar.*

ANS (see also Fig. 10): To create a heatmap showing insect activity based on the circadian rhythm: The y-axis will represent the hours of the day (0-23 hours). The x-axis will represent the dates or timestamps. The colour intensity will represent the insect count, with a colorbar indicating the correspondence between colour intensity and insect count. I'll proceed to create this heatmap for you.

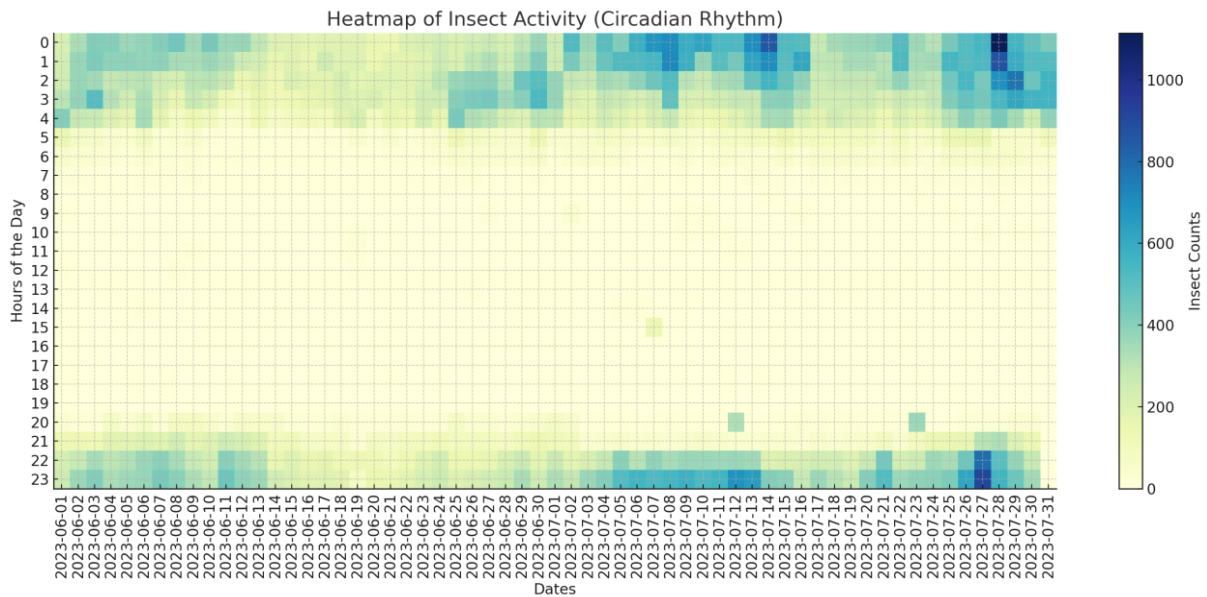

Figure 10. A heatmap of insect counts for 3 months. Lepidoptera like *H. armigera* are nocturnal getting a timestamp between 9 pm-4 am (*Tuta absoluta* is an exception and is responding to pheromones between 5-10 am). This information is important if we need to spray.

**Complex Queries**

*CQ1: How many unique trap locations (Latitude, Longitude) are there? Can you show them on a world map?*

NOTE: Due to nondisclosure of data, for this specific picture we have added some noise in the database's GPS coordinates.

ANS (see also Fig. 11): To answer this question, we'll follow these steps: Drop duplicates based on 'Latitude' and 'Longitude' columns to get unique trap locations. Count the number of unique locations. Plot these unique locations on a world map.

To determine the number of unique trap locations and visualize them on a map:

**Extract Unique Locations**:

Extract unique pairs of latitude and longitude from the dataset to identify unique trap locations.

**Visualization on a Map**:

Use a map visualization library to plot these unique locations on a world map.

Let's start by finding the number of unique locations and then visualize them on a map.

*CQ2: How do counts vary with latitude and longitude in Greece only? Create a heatmap overlayed on a Greek map. Use folium.*

NOTE: Due to nondisclosure of data for this specific picture, we have added some noise in the database's GPS coordinates.

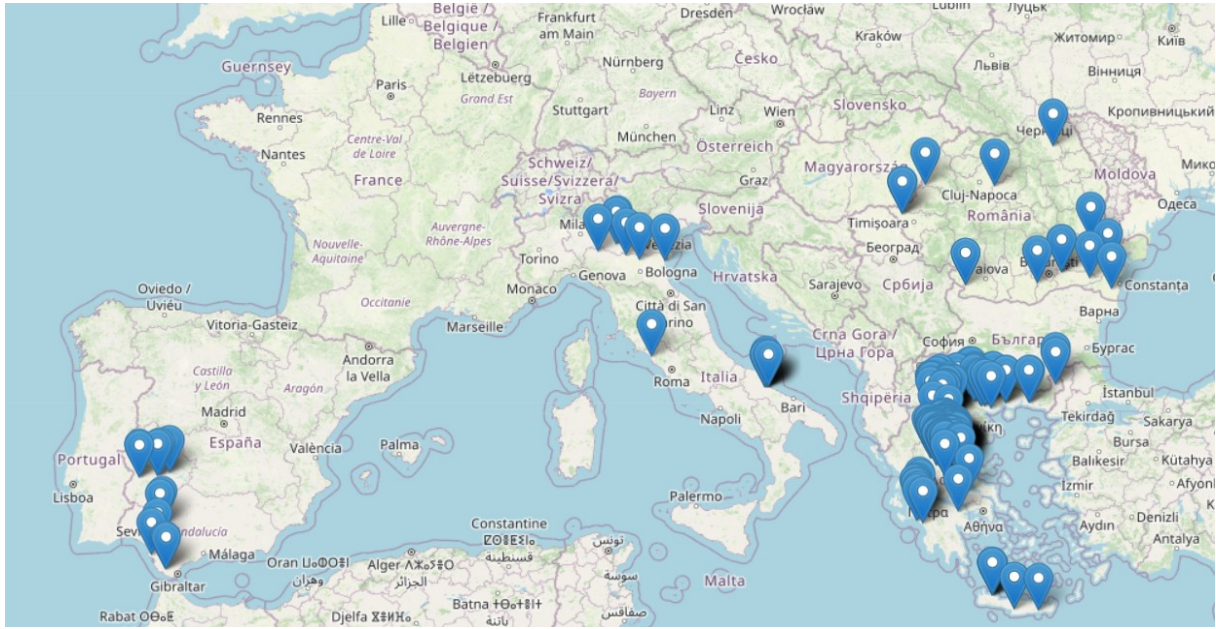

Figure 11. ChatGPT was asked about the location of the traps and to overlay them on a world map. Th stakeholders can immediately get a complex graph with a verbal request.

ANS (see also Fig. 12): To create a heatmap overlayed on a map of Greece, we'll need to: Filter the dataset to only include data points within the latitude and longitude bounds of Greece. Create a base map of Greece using folium. Overlay a heatmap using the HeatMap plugin, where the intensity is based on the insect counts.

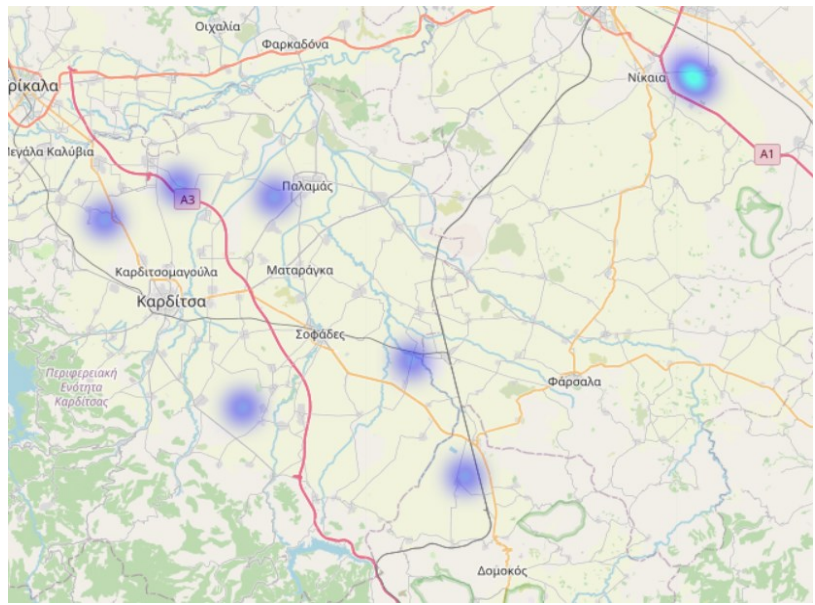

Figure 12. ChatGPT was asked about the insect pressure in Greece and to make a heatmap overlaid on a geospatial map. Th stakeholders can immediately see where the insect pressure is located (blue contours). Note that because e-traps report four times a day, this heatmap becomes animated if we stack daily heatmaps.

*CQ3: What is the average humidity and temperature and their standard deviation for the traps in Italy that captured weekly more than 100 insects in a week?*
ANS: To answer this question, we'll follow these steps: Filter the dataset to only include data points within the approximate latitude and longitude bounds of Italy. Group the data by trap ('Name') and week to compute weekly insect counts for each trap. Filter the grouped data to only consider entries where the weekly count of insects is more than 100. Compute the average and standard deviation of the humidity and temperature for these filtered entries.

```
The average humidity and temperature and their standard deviation for the
traps in Italy that captured more than 100 insects weekly are as follows:
Average Humidity: ≈61.29%
Standard Deviation of Humidity: ≈19.66%
Average Temperature: ≈26.46∘C
Standard Deviation of Temperature: ≈7.89∘C
```

*CQ4: Create a list of events for the trap with ID 213 based on the timestamp between 21.00pm and 4am and counts that you believe are outliers. Focus on outliers with high values. Each hour should be treated independently. Plot the timeseries and mark on it the outlier values.*

ANS (see also Fig. 13): To identify high-value outliers for each hour independently, we can employ the following approach:

**Data Filtering**: Select the data corresponding to trap ID 213, and filter it based on the specified hours. **Hourly Outlier Identification**: Group the data by the hour of the timestamp.

For each hour, calculate the mean and standard deviation of the counts. Identify outliers for each hour as values that are beyond a certain number of standard deviations from the hourly mean. We'll consider values more than 33 standard deviations away from the mean for each hour as outliers. **Create Event List**: Compile a list of events containing the timestamps and insect counts identified as outliers. These events represent timestamps where the insect counts were unusually high for their respective hours, being more than 3 standard deviations away from the mean count of that hour.

```
RESULT
             Timestamp   Counts
0   2023-07-30 21:00:00     8.0
1   2023-08-09 21:00:00     8.0
2   2023-08-10 21:00:00    12.0
3   2023-08-10 22:00:00    16.0
……….
17  2023-08-17 04:00:00     5.0
18  2023-08-25 04:00:00     5.0
```

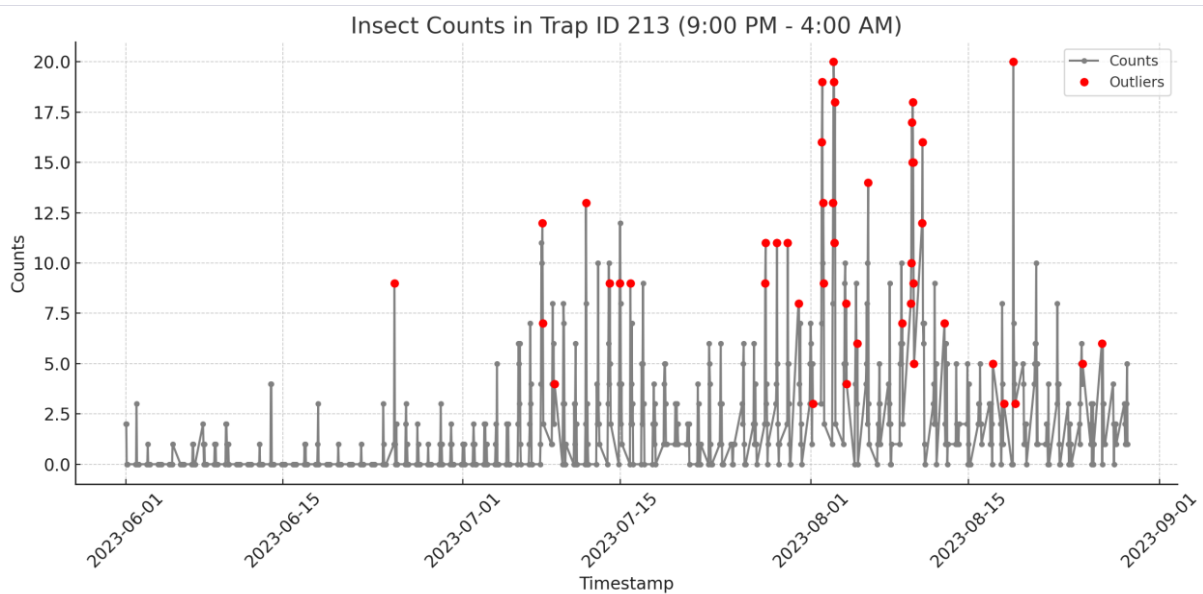

Figure 13. A timeseries of the insect counts of a specific trap with outliers superimposed.

*CQ5:* Determine whether the traps with IDs 213 and 217 have similar catch patterns.
ANS: (see also Fig. 14). To determine whether traps with IDs 213 and 217 have similar catch patterns, we can perform various analyses, such as:
**Visual Analysis:**
Plotting the timeseries data of both traps to visually inspect the similarities and differences in catch patterns.
**Statistical Analysis:**
We can calculate the correlation between the counts of the two traps to quantify the similarity in their patterns.
Performing a hypothesis test, like the two-sample t-test, to determine if there's a significant difference between the two traps' counts.
**Frequency Analysis:**
Analyzing the frequency or periodicity of the catches in both traps, which can be done using Fourier Transform to identify common patterns or cycles.
Let's start with the visual analysis and then proceed to statistical analysis. Shall we?
Visual Analysis
First, let's plot the timeseries of insect counts for traps 213 and 217 to visually inspect their patterns.
Statistical Analysis
After the visual analysis, we'll calculate the correlation between the insect counts in the two traps, which will give us a numeric value indicating how similarly the traps' counts move together. A correlation close to 1 indicates a strong positive relationship, while a correlation close to -1 indicates a strong negative relationship. A correlation around 0 indicates no linear relationship.

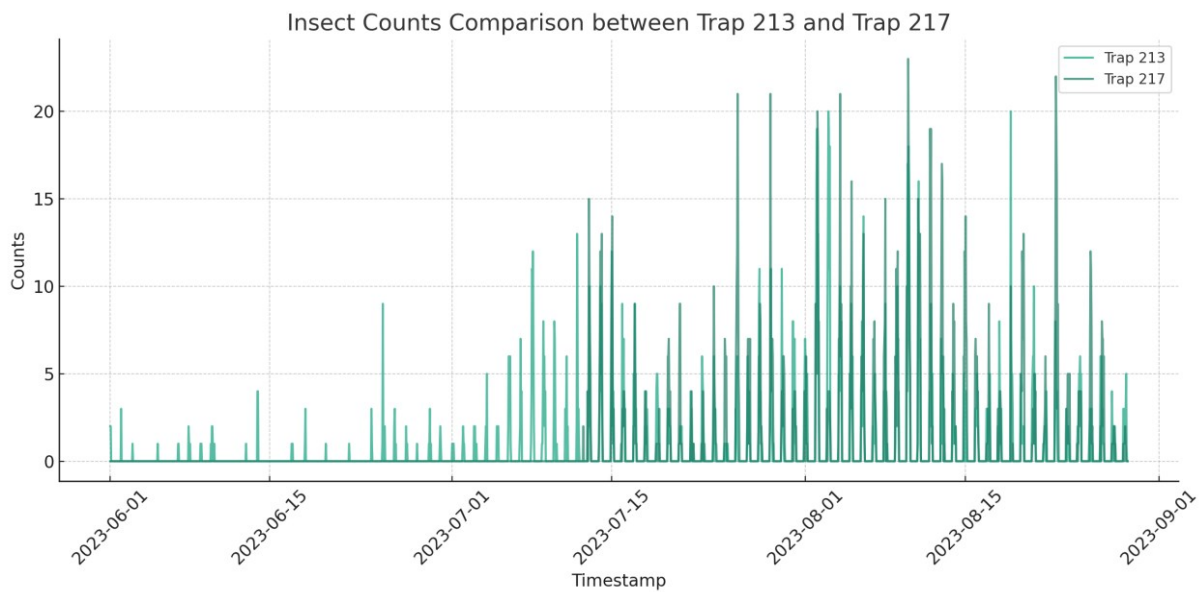

Figure 14. The plot above compares the insect counts between traps 213 and 217 over time. Each line represents the counts from one trap, allowing for a visual comparison of their patterns.

**Conclusion**
**Visual Analysis:** Visually, there are similarities in the patterns of insect counts between the two traps, as seen in the plot.
**Statistical Analysis:** The two-sample t-test did not show a statistically significant difference in the means of insect counts between the two traps.

Frequency Analysis: The FFT plot suggests that there are some common frequency components in the insect counts of traps 213 and 217, indicating similar patterns or cycles. Together, these analyses suggest that traps 213 and 217 exhibit similar catch patterns, corroborating the moderate correlation observed earlier.

*CQ6: Show a heatmap of Humidity and another one for temperature where the y-axis is the hours of the day and x-axis the timestamps. Visualize and present a colorbar that ranges from 0-100 for humidity and 0-60 for temperature.*

ANS (see also Fig. 15): Certainly! I'll create two heatmaps: Humidity Heatmap: Y-axis: Hours of the day (0-23 hours) X-axis: Dates Color intensity corresponds to humidity levels, ranging from 0 to 100%.

Temperature Heatmap: Y-axis: Hours of the day (0-23 hours). X-axis: Dates. Color intensity corresponds to temperature, ranging from 0 to 60°C. Let me create and display these heatmaps for you.

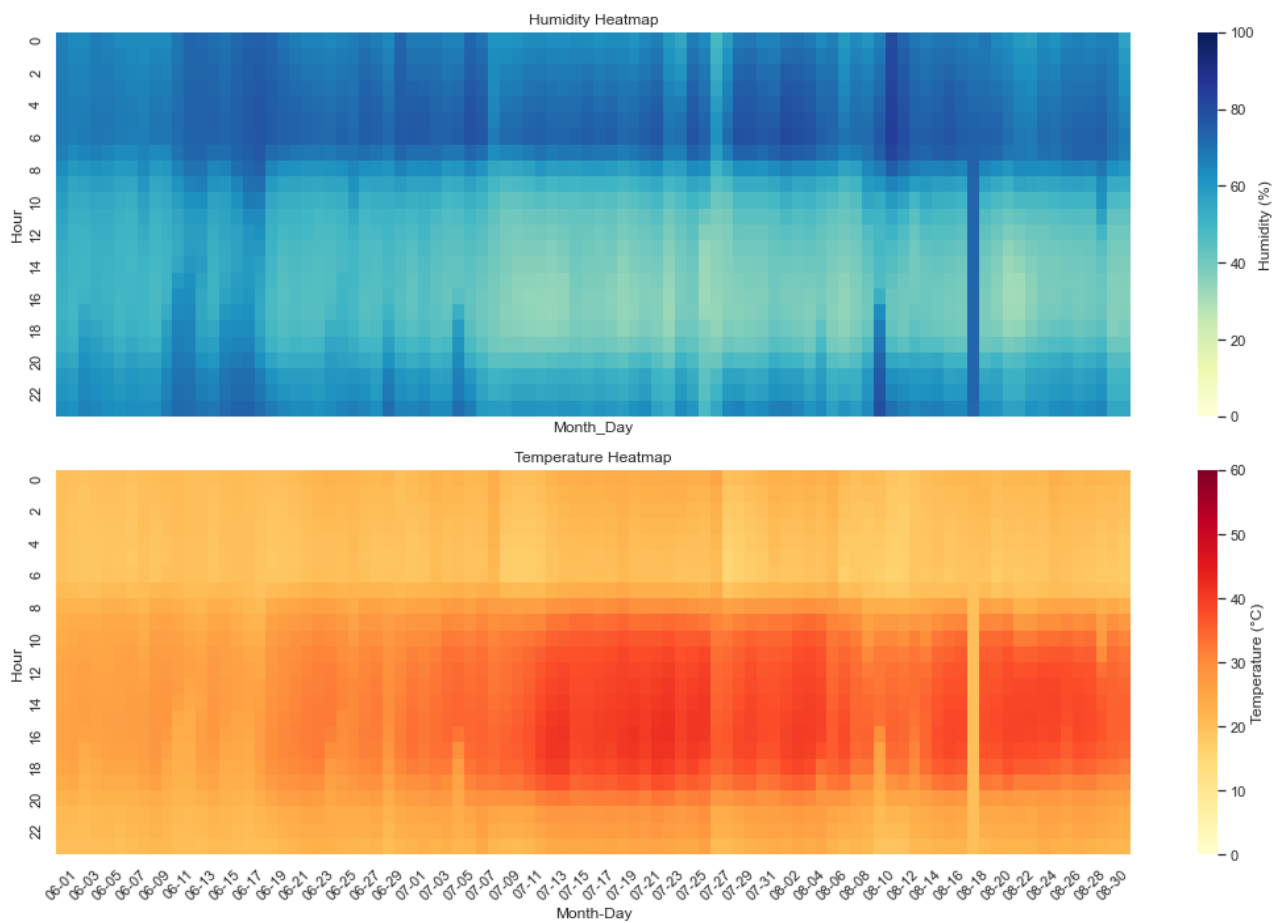

Figure 15. A representation of the microclimate around the e-traps. Hourly measurement (y-axis) of humidity (top) and temperature (bottom) for the whole monitoring period. It allows one to see the big picture of heat progressively increasing from June to August, identify several extreme heat events that occurred during the summer of 2023 in Southern Europe and identify the humidity variation patterns between day and night.

*CQ7. Find the closest three traps from the centre of the town Larissa. Perform cross-correlation analysis on their counts.*

NOTE: The database does not include the names of cities and their association to GPS coordinates. The LLM infers the connection itself from Latitude and Longitude measurements.
ANS: To perform cross-correlation analysis on the counts from the three closest traps to the centre of Larissa, we need to follow these steps: Find the Coordinates of Larissa's Centre:

The geographical coordinates of the centre of Larissa, Greece, are approximately 39.6396N latitude and 22.4196E longitude. Identify the Three Closest Traps: Calculate the distance of each trap to the centre of Larissa and identify the three closest traps. Perform Cross-Correlation Analysis: Conduct a cross-correlation analysis on the insect counts from the three identified traps.

```
RESULT
         Name       Lat       Long  Distance_to_Larissa
191260    198  39.638391  22.384232            3.039020
208694    206  39.609993  22.447136            4.049095
45812     127  39.684589  22.424285            5.011198

CROSS-CORRELATION RESULTS
Name       127       198       206
Name
127   1.000000  0.531957  0.296066
198   0.531957  1.000000  0.328235
206   0.296066  0.328235  1.000000
```

*CQ8. Is there a certain temperature and humidity threshold where the number of moths increases or decreases significantly?*

ANS (see also Fig. 16): To determine if there's a specific temperature and humidity threshold where the number of moths significantly increases or decreases, we can:

**Divide the Data into Bins**: Create temperature and humidity bins to categorize the data. For example, temperature bins could be <10°C, 10−20°C, 20−30°C, etc., and humidity bins could be in increments of 10%.

**Calculate Average Counts**: Calculate the average count of moths for each bin. Visualize the Results: Create a visualization to easily identify where significant increases or decreases occur.

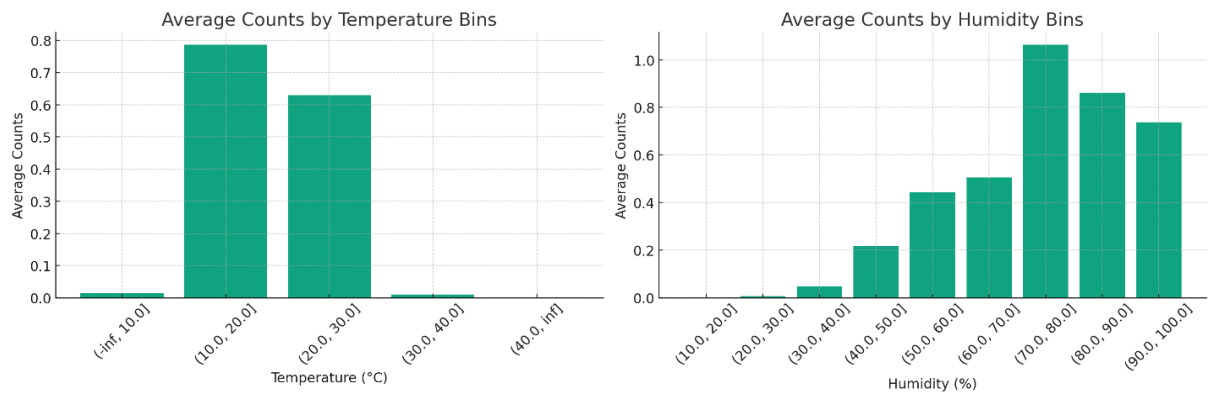

Figure 16. Average counts of insect capture w.r.t temperature (Left) and humidity (Right) intervals.

*Temperature*

**Low Temperatures (< 10°C):** The counts are relatively low, indicating less activity in colder conditions.

**Moderate Temperatures (10-20°C and 20-30°C):** The counts are higher in these ranges, showing that moderate temperatures are more favorable for the moths.

**High Temperatures (> 30°C):** The counts decrease significantly in very high temperatures, indicating that extremely hot conditions are not suitable for the moths.

*Humidity*

**Low Humidity (< 40%):** The counts are quite low, suggesting that dryer conditions are not favorable.
**Moderate to High Humidity (40-90%):** The counts increase and are relatively higher in these humidity ranges, indicating that these conditions are more suitable.
**Very High Humidity (> 90%):** There's a noticeable decrease in counts, suggesting that extremely high humidity levels might not be as favorable.
From these observations, it seems that the moths are more active under moderate temperature and humidity conditions. Extreme temperatures and either very low or very high humidity levels seem less favorable for *H.armigera*.

*CQ9. Peak Activity: During which hours are the insects most active, and how does this relate to the prevailing temperature and humidity conditions?*
ANS (see also Fig. 17): **Hourly Activity Analysis**: Calculate the average number of moths counted each hour to determine their most active hours.
**Temperature and Humidity Analysis**: Calculate the average temperature and humidity each hour to understand the prevailing conditions. Comparison: Compare the hourly activity of moths with the hourly temperature and humidity to identify any patterns or correlations. Let's start by extracting the hour from the Timestamp and then proceed with the analysis.

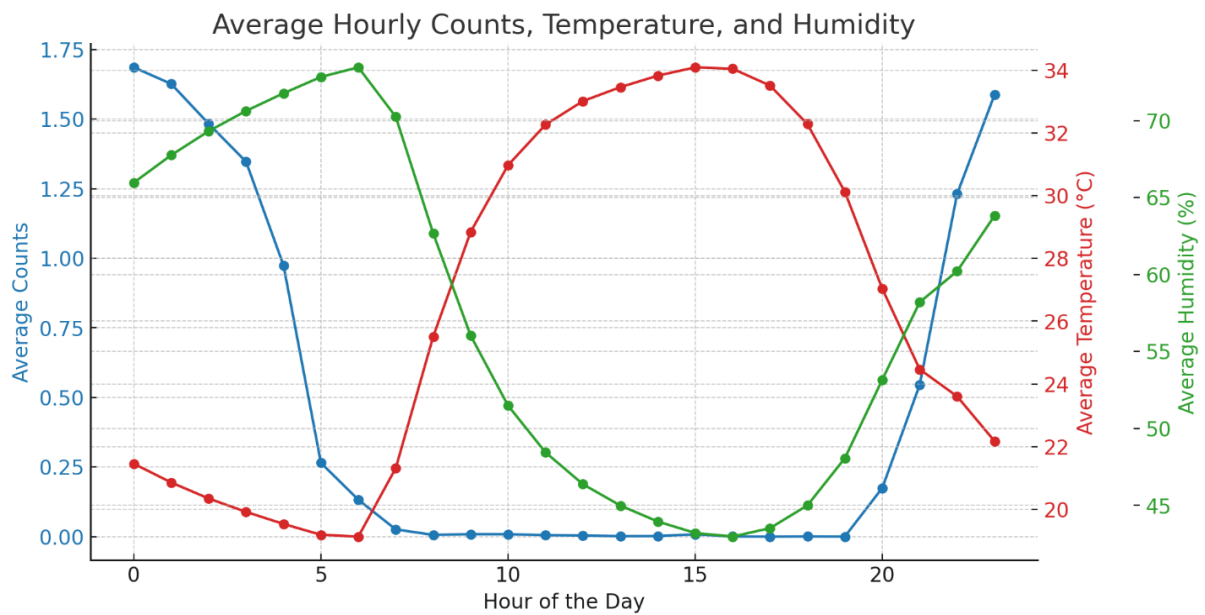

Figure 17. Average counts of insect capture per hour (y-left axis) superimposed on temperature and humidity diagrams.

The plot above shows the average hourly activity of moths (blue), along with the average hourly temperature (red) and humidity (green) levels. Average Counts (Blue Line):
The moths seem to be most active during the early hours of the morning, peaking around 1-3 AM (Note: this is *H. armigera*).
There is another smaller peak in activity in the evening, around 7-8 PM. Average Temperature (Red Line): The temperature tends to be lower during the early hours when the moth activity is highest. The temperature starts increasing from the morning and peaks around 3 PM, when the moth activity is quite low. Average Humidity (Green Line): The humidity is higher during the early hours of the morning, coinciding with the peak in moth activity. The humidity decreases as the day progresses and is at its lowest in the afternoon, aligning with the lower moth activity. These observations suggest that *H. armigera* moths tend to be more active during cooler and more humid parts of the day, primarily during the early morning and evening hours.

**The TreeVibe probe monitoring woodboring insects**

**Queries**
*Q1: Fetch the latest vibration recording from the sensors in Heraklion, Crete and show me a) the waveform in time, b) the power spectral density and, c) the spectrogram.*
ANS (see also Fig. 18): Certainly! Here's a breakdown of the steps we will follow:
**Waveform in Time Domain**: Load the audio file and extract the audio signal and sampling rate. Plot the audio signal against time to visualize the waveform.
**Power Spectral Density (PSD)**: Estimate the power spectral density of the audio signal to show how the power of the signal is distributed over different frequencies. Plot the estimated power spectral density against frequency.
**Spectrogram**: Compute the spectrogram of the audio signal to show how the frequency content of the signal varies over time. Display the spectrogram as an image where the x-axis represents time, the y-axis represents frequency, and the color represents the magnitude of the signal.

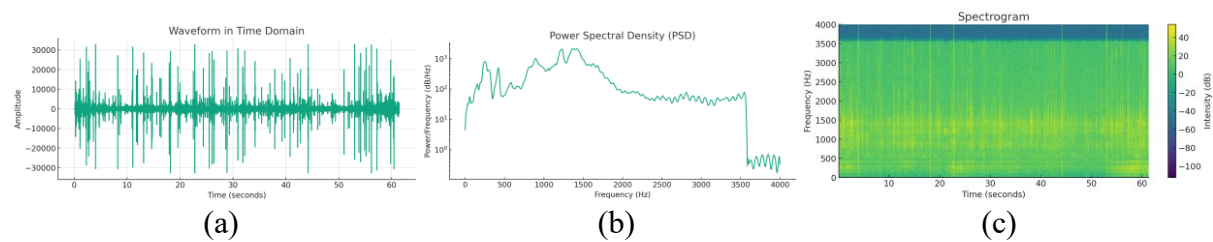

(a)  (b)  (c)

Figure 18. A vibrational recording from an infested tree. The pulse train seen on (a) is characteristic of fibre cracking as the larvae eat and move in the tree. The frequency content (PSD) in (b) and the spectrogram in (c) give us valuable information on the frequency content of the cracks that appear as vertical stripes in the time-frequency domain on the right because impulses are short time and broadband signals.

The Figures 18 a-c are often used by entomologists to determine the infestation status of trees [69-70].
However, when we ask the more difficult question of inferring if the waveform originates from an infested tree, though the tactic it presents is correct, its decision was inconclusive. A human observer would immediately see the pulse trains in the time domain (Fig. 18a) that translate into vertical stripes in the time-frequency domain (Fig. 18c) and would immediately classify it as infested. The tree is indeed infested. The answer of the ChatGPT is as below:
"To determine if the tree might be infested based on the recording, we can analyse the frequency content of the recording. Insects, such as wood-boring beetles, may produce specific sounds (e.g., dew to chewing, tunnelling) that fall within certain frequency ranges. Here's the plan: Load and visualize the audio waveform. Compute the spectrogram to display the frequency content over time. Analyse the dominant frequencies and their patterns. By examining the spectrogram and dominant frequencies, we can infer if there are consistent sounds that might indicate the presence of insects…..Given the data at hand, while there is a dominant frequency, we cannot conclusively determine infestation without additional context or comparisons."

**The WingBeats recorder**
**Queries**

Q1: This is a recording of a mosquito. Write a Python program that plots a) the power spectral density and, b) the spectrogram.
ANS (see also Fig. 19): The answer is the same as in the TreeVibe example.

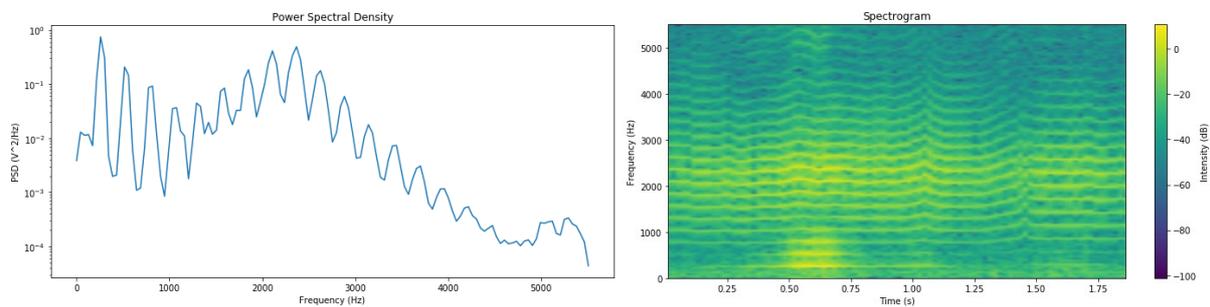

Figure 19. An optical recording of the wingbeat of a mosquito. The PSD on the left shows us the wing beating frequency (the first lobe) and the harmonics that are characteristic of this species, The spectrogram corroborates this information.

When we ask ChatGPT if it can discern the sex of the mosquito from its sound, again, it proposes the correct strategy, but it fails to conclude correctly. Although this question may sound difficult to a reader, it is generally not because mosquitoes are dimorphic and the females are larger and, therefore, their wingbeat frequency is lower than males.

ANS: To determine the gender of the mosquito based on its wingbeat frequency, we'll analyse the frequency content of the recording. Typically, female mosquitoes have a lower wingbeat frequency than males. Here's a general guideline: Female mosquitoes: approximately 250-500 Hz. Male mosquitoes: approximately 500-800 Hz. Let's proceed to analyse the frequency content of the recording you provided……We can't conclude.

**The vision-based insect trap**

ChatGPT can effectively find and retrieve requested images from the server using the speech modality. Note that the traps themselves can be dispersed in various parts of the world but through wireless communication they establish internet connection and report to the same server. It can efficiently apply signal processing techniques and present the Python code.

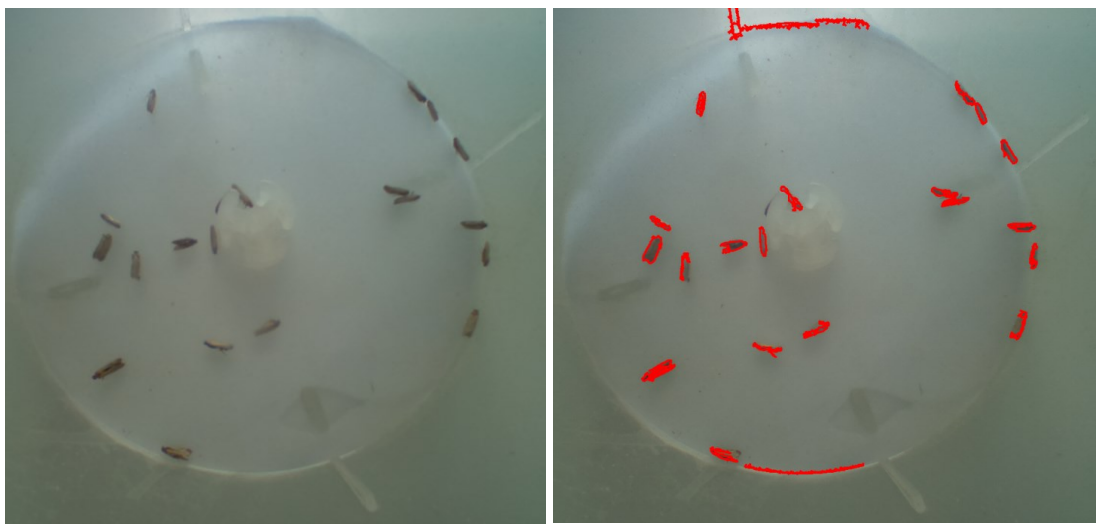

Figure 20 [best viewed in colour]. (Left) *Plodia interpunctella* 20 insects. (Right) Processed image by ChatGPT counting 22 insects.

However, as the following interaction shows after presenting it Fig.20-left, it can apply signal processing techniques (see Fig. 20-right) but it cannot count efficiently or recognize insects yet and it explains the reason why. It makes consistent mistakes in counting.

Q1: Can you count the insects in the photo?

ANS: I'll start by loading the image to visualize its content. After that, I can attempt to count the insects in the photo, depending on the clarity and quality of the image. Let's take a look. I do not have an embedded object detection model within this environment. However, I can use traditional computer vision techniques, as demonstrated earlier, to attempt to identify and count objects in an image. For more sophisticated object detection tasks, pre-trained models like YOLO, SSD, or Faster R-CNN are usually employed. These models are trained on large datasets and can detect multiple object classes with high accuracy.

**Discussion**

ChatGPT can transcribe accurately voice commands related to data analytics to text. The quality of synthesized voice that uses to respond (i.e., text to speech synthesis) is sufficient for general applications. The prompts that correspond to voice commands must not be vague. Commands that are related to direct applications of signal processing techniques and especially to informational retrieval on tabular data are carried out correctly at high precision. Queries that relate to interpretating the content of images, video, and audio (i.e., content interpretation of corpora) in the context of insect, plant, and disease recognition in plants and trees, have various degrees of accuracy but generally are inconsistent and inaccurate yet. ChatGPT lags behind two issues a) Incorporation of generic and specific object detection, and b) content interpretation and inference using principled and transparent methods. The plugins and their associated models need to be adapted for the case of insects and plant diseases. Note that current technology supports models that are being very accurate in dealing with these questions [46-49], but these models are not yet incorporated to the ChatGPT's plugins.

Although in CQ8, CQ9 and several others attempts to decide based on the data, it generally lacks principled ways to interpretate image and audio content. If the above limitation are surpassed then it will be able to fuse heterogeneous information from diverse sensing modalities to infer and abstract the situation in the agricultural field (see Section Queries from the future section below).

The user must always examine the tactic (i.e., the algorithmic approach) that the LLM intends to follow and ask for code implementation (Python in our case) for human evaluation. We are far for blindly trusting the code for critical decision making about data analytics as applied to precision agriculture.

Many of the findings in Figures 5-19 regarding circadian rhythms, pheromone responses and the effect of temperature and humidity on insect catches can be found in the bibliography. However, in our case, we can derive them directly from the field, quickly, precisely, and effortlessly. We envisage a continuous stream of information flowing from thousands of nodes placed in agricultural crops and trees across various countries and basins. Leveraging IoT communications, this data is transmitted to a cloud-hosted service. The service then abstracts this information, presenting it in the form of accessible graph networks. Stakeholders can interact with these graphs, querying and visualizing them using natural, unconstrained speech through Language Learning Models (LLMs). ChatGPT can be asked to engage in a discussion on the nature of the data it examines and by keeping track of previous queries and answers, provide refining analysis and elaborating visualizations on demand. This is not the way data analytics in precision agriculture is currently performed, therefore, we suggest that LLMs can become disruptive in this domain.

The concept of introducing ChatGPT in the pipeline of data analytics makes this service responsive and accessible to the general audience like entomologists, biologists, agronomists that are not always inclined towards computer science thus significantly increasing their productivity.

**Future prospect**

If the image processing capabilities of ChatGPT evolve so that it can examine the content of images related to pests, plant diseases and crops reliably, as well as audio from microphone and vibrational sensors then it would be able to process image data from drones and tractors to monitor crop health and detect diseases or pest infestations early. Currently, LLM rely on signal processing techniques that have not been adapted for the special case of insects, diseases, and plants and even generic object detection algorithms are not immediately available. However, this adaptation to the specific domain of insect data, plant diseases, plants and trees is feasible using current technology and has been reported to achieve great accuracy [50-64]. Once the object detection algorithms have been adapted to these domains, a human observer then could pose very specific questions such as:

**Queries from the future**
**Vision based traps:**
*Q1: Can you count the male and female Bactrocera oleae insects in the provided picture? A human can discern this based on the ovipositor present only in the females and the abdominal shape.*

*Q2: Can you count how many Drosophila melanogaster and how many Drosophila suzukii are in the same picture? Drosophila suzukii has distinctive dark spots at the tips of its wings, whereas Drosophila melanogaster typically has clear wings.* These questions are important to the producers as Drosophila suzukii is a pest whereas Drosophila melanogaster is not and both can be present in a fruit-fly trap.

Q3: There are thousands of automated camera traps for biodiversity assessment photographing animals in the wild that usually store the picture in embedded SD cards from where the images can be retrieved. If they have a communication ability to upload their images to a server then ChatGPT can retrieve and process them on demand. Currently, ChatGPT cannot access the Megadetector's animal recognition deep learning architecture [65-68] and the plugins it currently has (cannot analyse animal images efficiently.

**Vibration based probes:**
*Q1: Analyse the frequency content of this recordings <submit recording>. Is the tree infested with woodboring insects or not?*
*Q2: Analyse the frequency content of these vibrational recordings originating from infested mulberries and infer which pest it is.*
*Q3: Cross-correlate the results from 4 different trees and infer the probability that the plantation is infested.*
*Q4: Estimate insect density of an infested tree trunk.*
*Q5: Triangulate the vibrational recordings from many probes on the same tree and give a rough estimation of the location of the larvae in the trunk.*
*Q6:* Microphones for the monitoring of biodiversity [69-71]. *Given specific GPS coordinates and the species list of local birds and bats fauna recognise the vocalizing species.*

**Optical counters**
Optical counters can be naturally inserted in fruit fly, lepidoptera and mosquito traps. Here are some complex user-cases that a human operator can ask through the ChatGPT interface:

C1: Perform kriging and identify hotspots of insect activity. ChatGPT could control and coordinate autonomous agricultural machinery, such as drones, tractors, and irrigation systems, based on real-time data and predefined protocols working on a 24/7 basis. The spraying drones can be guided for precision interventions at the hotspots [72]. Re-estimate the situation after intervention and repeat the process.

C2: Arrays of optical sensors that register the backscattered infrared light from insects' wingbeat are mounted on moving platforms such as tractors and drones and scan the agricultural fields (see Fig. 21).
All data can be interfaced via the ChatGPT that asks refining questions on insect density, hotspots and insect species recognition based on the registered wingbeat.

C3: ChatGPT can integrate optical counters in the entrance of the beehive as well as environmental and vibrational data [12-13] and monitor a dispersed network of colonies. It could ask direct questions like: What is the health status of the beehive? Are the bees going to swarm? Is it the right time to collect honey? Do you sense severe presence of parasites in the beehive?

C4: Optical counters utilizing backscattered light in mosquito traps: From the counted number of mosquitoes sucked in the trap with respect to time, derive their activity and circadian rhythm so that you plan intervention accordingly (insecticides are effective when they encounter mosquitoes in flight). How many of the trapped mosquitoes are female? This is an important factor in the sterile insect technique. How many of the females are gravid? (use near infrared light sensors embedded in traps). Connect with barcoding apparatus and decide if a batch of trapped mosquitoes is infected with west Nile virus. Given specific GPS coordinates and temperature (insects are ectothermic and their wingbeat changes and need to be considered) and the list of local mosquito species (insect fauna is vastly varying around the globe), recognise the mosquitoes from their wingbeat up to genera and species levels.

C5: ChatGPT be connected to the reporting database of a dispersed network of airborne spore sampling and analysis systems operating in the field [73]. Devices suck in air, isolate spores, perform *in-situ* isothermal amplification of DNA and the results are transmitted to a server. ChatGPT can retrieve data and combine them with weather data to provide an indication of the risk of infection, improving spray decision-making.

C6: LLM can naturally cooperate with Decision Support Systems as the latter boil down to a chain of if-then-else reasoning based on thresholding sensors' measurements and integrating the timestamps [74]. They can cooperate with ChatGPT to receive input from farmers and provide back personalized recommendations and actionable insights, considering specific crop types, local conditions, and individual preferences communicated using synthesized speech.

C7: E-trap insect captures can provide direct input to population density models to initialize them so that they infer the population in the field. Again, ChatGPT can moderate the interaction between the human operator and the database in a more responsive and intuitive way that the currently manual approach [75].

C8: Lastly, ChatGPT can gather and analyse real-time market data, helping farmers make decisions about when to sell their crops and at what price, potentially increasing profitability and accepting the usefulness of precision agriculture in common practices [76].

Prompts play a vital role in effectively utilizing large language models since they facilitate communication between users and the model. Crafting well-designed prompts is essential, as they determine the quality of responses. This requires an understanding of the model's capabilities and limitations, as well as a clear understanding of the specific task or application.

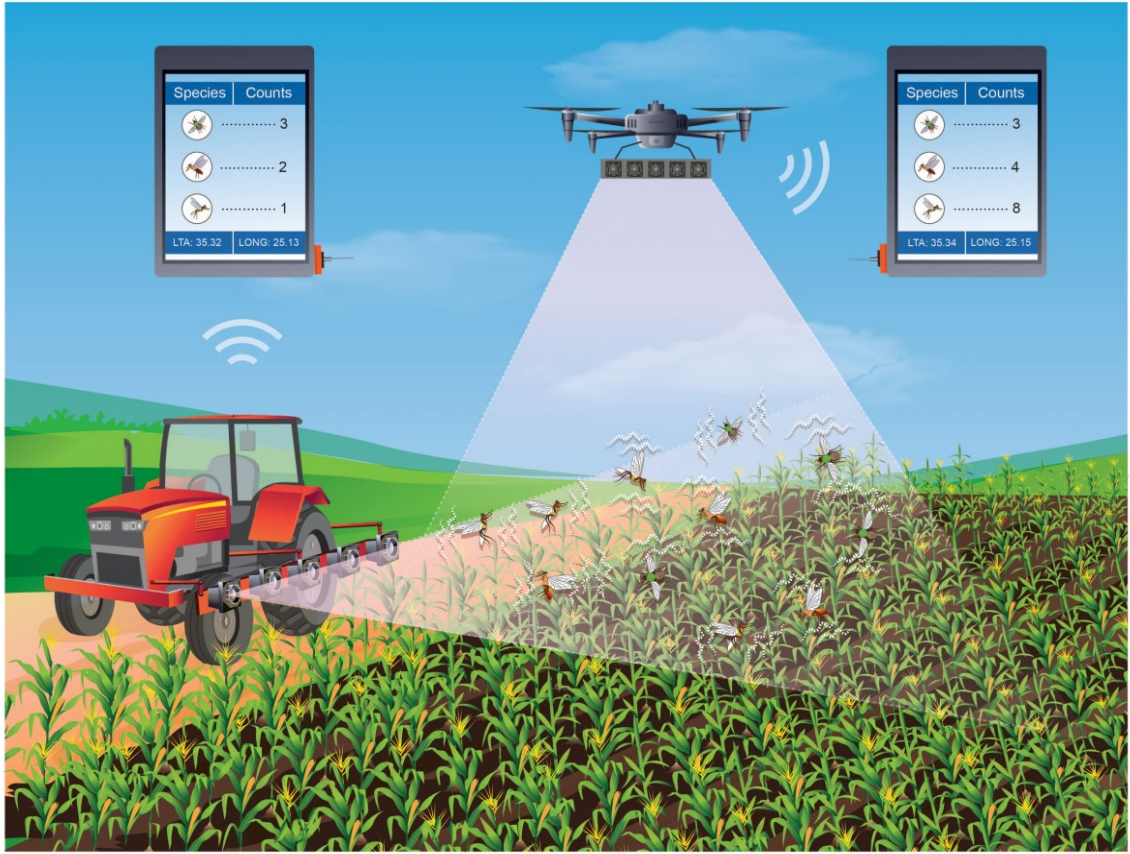

Figure 21. Linear arrays of optical counters and wingbeat recognizer use backscattered infrared light to scan fields and report insect density and insect species composition of flying insects. ChatGPT can help in the fusion of heterogeneous information and information retrieval which is an import part of decision making.

Effective prompts can lead to more accurate and contextually relevant responses. During the process of composing queries in the Results Section, we found it necessary to avoid vague or poorly constructed prompts that could result in less useful or confusing outputs. To improve the quality of generated Python code, it is advisable to restate the question while providing as much information as possible. Additionally, explicitly specifying the use of Python libraries, such as GeoPandas and Folium can enhance the accuracy and relevance of the generated Python code. ChatGPT still lacks the connection with general object detectors and adapted models for the special case of entomology and agriculture. This connection is currently technologically feasible. E-traps and sensing probes needs to improve their accuracy and sensitivity respectively but at a lower cost for hardware (i.e., electronics and IoT services, cloud services) and consumables (e.g., pheromones, scents, chemicals for DNA analysis etc), so that are widely accepted and applied at global scales.

## Conclusions

The integration of ChatGPT as an interaction frontend that directly accesses and manipulates the database of servers containing heterogeneous data in the context of precision agriculture offers substantial benefits. Asking questions in an unconstrained verbal fashion about real-time data eliminates the substantial technical expertise typically required for processing and interpreting data from traps and sensors in fields and trees. The immediate response and the visualized results the operators get, the ability of LLMs to engage in discussion through iterative and refining questions enhances the precision, efficiency, and sustainability of pest management strategies, contributing to better agricultural practices, reduced chemical usage, and improved overall crop health.

The current level of speech recognition accuracy enables practically flawless interaction, as the vocabulary of mathematical and data processing terms is limited. While diverse dialects and accents could potentially affect the accuracy of speech recognition systems, it is important to note that this work does not address in-field applications that require operating booths capable of withstanding harsh outdoor conditions, including noise and extreme weather conditions, to ensure reliable operation. During this research, we interacted with the remote server's database through ChatGPT in controlled and calm conditions.

Large Language Models (LLMs) have the capability to integrate information from various sources, such as data from soil measurements, moisture levels, satellite imagery and weather forecasts. This integrated data can be queried to optimize the utilization of water, fertilizers, and pesticides, effectively reducing waste, and minimizing environmental impact. Additionally, this approach can help reduce the cumulative economic impact of pests on agricultural operations.

Integrating ChatGPT into existing data pipelines is an intuitive process. However, in our opinion, it is crucial to consistently demonstrate the Python code for GPT-4 (in LangChain set verbose=True). This approach ensures that users can crosscheck whether their questions have been accurately translated into Python code. Human operators may not always convey messages accurately, so users should not blindly rely on ChatGPT's outputs when making decisions based on data. ChatGPT still misses plugins that will function as tools for inference on specific domains that fetch a result to the LLM to be integrated with the textual message (i.e., direct access to object detection/recognition libraries of general and specialized content, Matlab/Mathematica platforms, software that requires API Keys etc.).

## Data Availability

Insectronics [31] is a private entity and does not disclose data. In this paper only collective results are presented and when deemed necessary, data has been perturbed.

## Acknowledgments

We thank Dr. Iraklis Rigakis of Insectronics for granting us restricted access to the database of the company from which we were allowed to draw and report only descriptive data.

## APPENDIX

In this section we disclose the code in Python provided by ChatGPT. All questions have been submitted using voice commands and have been recognized by 100% from the first try.

The code can be found at this address:https://github.com/potamitis123/ChatGPT-in-the-context-of-precision-agriculture-data-analytics (accessed on 25/10/23). The code can be applied in any tabular dataset of the form described in Table. 1.

Birds, Audio/Vibrations

Spores

Decision Support

Population models